\documentclass{article}

\usepackage[preprint]{neurips_2026}

\usepackage[utf8]{inputenc} 
\usepackage[T1]{fontenc}    
\usepackage{hyperref}       
\usepackage{url}            
\usepackage{capt-of}
\usepackage{booktabs}       
\usepackage{amsfonts}       
\usepackage{nicefrac}       
\usepackage{microtype}      
\usepackage{graphicx}
\usepackage{subcaption}
\usepackage{multirow}
\usepackage{colortbl}
\usepackage[table,xcdraw]{xcolor}
\usepackage{enumitem}
\usepackage{pifont}
\usepackage[most]{tcolorbox}
\usepackage{listings}
\lstdefinestyle{promptstyle}{
  basicstyle=\ttfamily\small,
  breaklines=true,
  columns=fullflexible,
  keepspaces=true,
  showstringspaces=false
}
\title{Tool-IQA: Augmenting Image Quality Assessment \\ with Simple Tools}

\author{
Guanyi Qin$^{1,2,\dagger}$, \
Junjie Zhang$^{3,\dagger}$, \
Chunming He$^{4}$, \
Yibing Fu$^{1}$\\
\textbf{Jie Liang}$^{2}$, \
\textbf{Tianhe Wu}$^{5}$, \
\textbf{Lei Zhang}$^{2,6,}$\thanks{Corresponding author. $^\dagger$Equal Contributions.}\\
$^1$National University of Singapore, $^2$OPPO Research Institute \\
$^3$Nanyang Technical University, $^4$Duke University \\
$^5$City University of Hong Kong, $^6$The Hong Kong Polytechnic University \\
	{\tt\small guanyi.qin@u.nus.edu} \\
    \url{https://github.com/narthchin/Tool-IQA}
}

\begin{document}

\maketitle

\begin{abstract}

Vision-Language Models (VLMs) have been increasingly adopted for Image Quality Assessment (IQA). However, current methods typically employ a static \textit{one-shot} scoring paradigm, despite the fact that humans assess image quality through dynamic visual inspection, \emph{e.g.}, selectively adjusting views to verify details and subtle artifacts.  Specifically, relying solely on a single-pass observation introduces two primary limitations: first, perceiving the image only at a global scale restricts the assessment of finer local details; second, the original intensity distribution of the image may overwhelm the visibility, leading to insufficient inspection of image quality. To address these issues, we propose \textbf{Tool-IQA}, shifting the assessment mechanism from passive scoring to a tool-augmented workflow. In particular, we equip VLMs with simple yet effective view tools: a \textit{Magnifier} to inspect local details, and a \textit{Gamma Corrector} to uncover visibility and hidden artifacts. The assessment follows a structured pipeline that consists of an initial observation with rubric notes, a tool-augmented in-depth inspection, and a final quantification for calibrated quality score. Furthermore, to ensure efficient and purposeful tool callings, we introduce a batch-aware training strategy to reward tool interactions that can yield positive contributions rather than simply encouraging usage. Experiments on a variety of IQA benchmarks demonstrate that, with effective tool calling and calibrated assessment, our proposed Tool-IQA significantly outperforms existing state-of-the-art models, \emph{e.g.}, it achieves a PLCC of 0.854 on the challenging CLIVE dataset.

\end{abstract}

\section{Introduction}
The goal of Image Quality Assessment (IQA) is to emulate the Human Visual System (HVS) in terms of perceptual quality~\cite{wang2004image}. Conventionally, this task was performed using handcrafted statistical features~\cite{mittal2012no,7094273,5705575,mittal2012making}, followed by the adoption of Convolutional Neural Networks (CNNs) and Vision Transformers (ViTs) to learn data-driven representations~\cite{6909620, ke2021musiq, Qin_Hu_Liu_Zheng_Liu_Li_Zhang_2023,9156687,Wang_Chan_Loy_2023,8576582}. With the advances of large Vision-Language Models (VLMs) \cite{qwen2025qwen25technicalreport}, which are pre-trained on massive data and possess profound image understanding capabilities, a significant transformation of IQA model training is underway~\cite{wu2024qalign,zhu2024compare2score}, not only in improving the scoring performance but also in the assessment patterns from scalar regression to more comprehensive, interpretable ones.

\begin{figure}[!t]
    \centering
    \includegraphics[width=\linewidth]{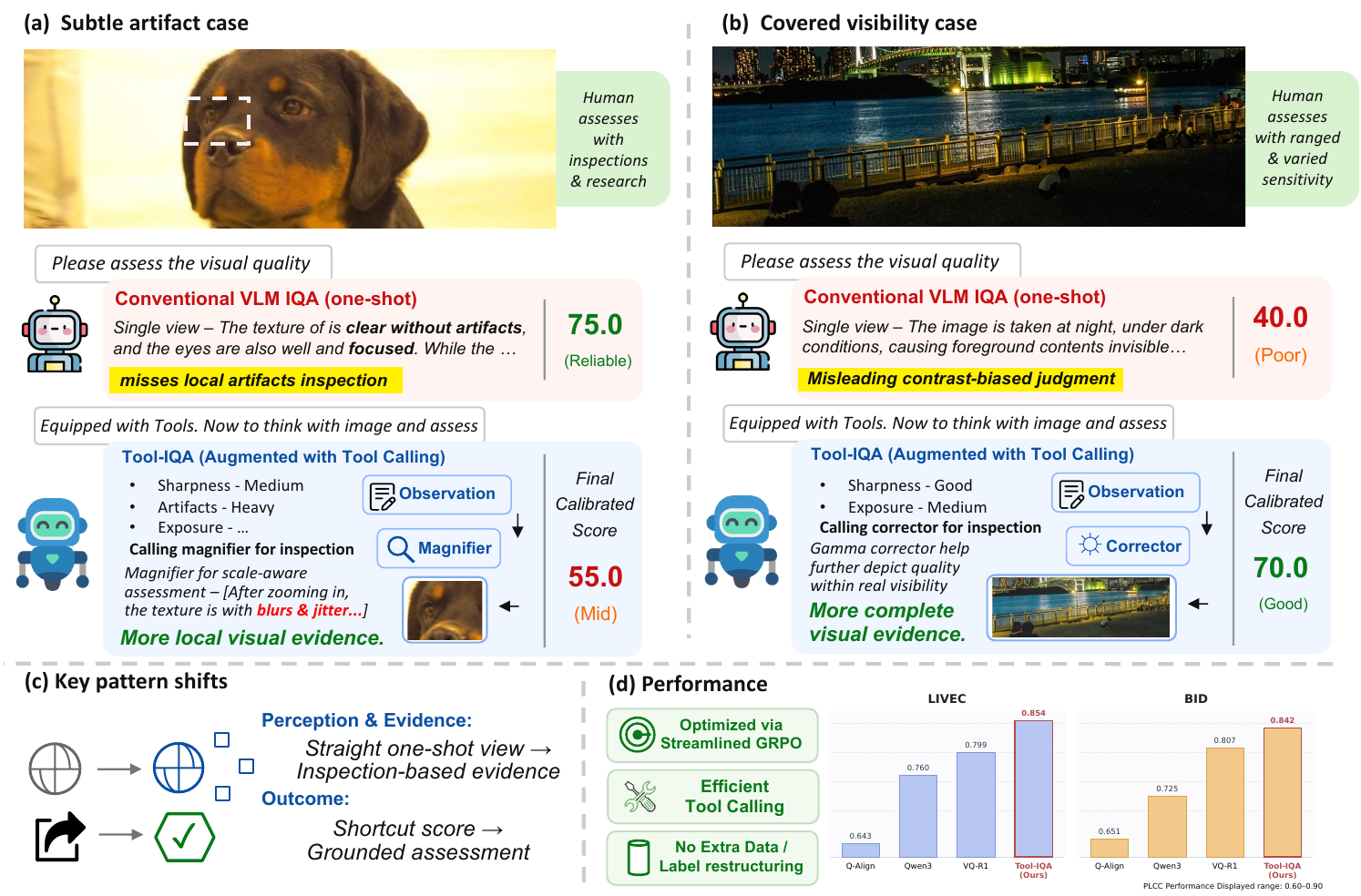}
    \vspace{-3mm}
    \caption{
    Existing VLM-based IQA methods typically maintain a one-shot pattern, which can lead to (a) \textit{scale-limited failures} by overlooking artifacts and (b) \textit{visibility-biased failures} by misattributing to poor quality. Subfigure (c) shows our proposed pattern shifts for IQA, and (d) shows that Tool-IQA leads to more satisfied performance by utilizing simple tools without extra data and annotation.}
    \label{fig:demo}
    \vspace{-5mm}
\end{figure}

To align VLMs with IQA purpose, strategies ranging from prompt engineering to Supervised Fine-Tuning (SFT) have been employed~\cite{10.1007/978-3-031-73232-4_14, wu2024qinstruct}. For example, Q-Align~\cite{wu2024qalign}  reformulates IQA as a text-defined rating task and Q-Instruct~\cite{wu2024qinstruct} constructs instruction-following data for quality-aware understanding, respectively. Inspired by recent RLHF breakthroughs~\cite{Guo_2025}, reinforcement learning has also been introduced to optimize VLM-based quality assessment beyond SFT. A representative work is VisualQuality-R1~\cite{wu2025visualqualityr}, where Group Relative Policy Optimization (GRPO) \cite{Guo_2025} with ranking rewards is employed to assess the multiple candidates sampled for each image-prompt pair and optimize the model according to human annotations. Such a reinforcement mechanism effectively improves scoring accuracy~\cite{li2025qinsight,wu2025visualqualityr}.

Notwithstanding these advances, a fundamental problem persists in the assessment workflow: determining image quality from a single static glance is inherently insufficient. HVS~\cite{Nassi2009,Najemnik2005} mitigates ambiguities through a sophisticated visual inspection process by exploring and inspection: observers selectively adjust their views to access local details~\cite{TREISMAN198097}, while adapting visual sensitivity to inspect subtle artifacts hidden under various lighting conditions~\cite{Carandini2012}. In contrast, current VLM-based frameworks mostly adopt a passive \textit{one-shot} workflow~\cite{Qin_Hu_Liu_Zheng_Liu_Li_Zhang_2023,wu2025visualqualityr} using a single-pass observation at a global scale and a fixed contrast.
This limitation manifests as \textit{scale-limited failure}, where subtle artifacts are overlooked or misidentified under the global scale, as shown in Fig.~\ref{fig:demo}(a); and \textit{visibility-biased failure}, where low-light or contrast-dominated scenes lead to misleading quality attribution, as shown in Fig.~\ref{fig:demo}(b). Consequently, without the affordance for in-depth perception, these models are prone to exploiting shortcut cues or illusions for quality assessment.

To bridge this gap, we propose to align VLM-based assessment models with the HVS inspection process, equipping them with the interactive mechanisms to perform sequential observation, \emph{e.g.}, observing the global context, characterizing regions that require closer and detailed inspection~\cite{zheng2025deepeyes}. To this end, we propose employing simple yet effective view tools to supplement the initial input, \emph{i.e.}, a \textbf{\textit{Magnifier}} for local inspection and a \textbf{\textit{Gamma Corrector}} for visibility-aware inspection to uncover more details, reformulating the assessment process from passive scoring to a tool-augmented workflow.
Based on these principles, we propose \textbf{Tool-IQA}, a novel tool-augmented framework optimized via GRPO \cite{Guo_2025}, where the inspection of complex visual attributes is explicitly externalized to and supplemented by auxiliary tools. It is worth mentioning that Tool-IQA achieves superior performance through effective adaptive perception, \textbf{without requiring additional datasets or restructuring existing annotations}, resulting in stronger capability to characterize evidence and produce calibrated assessments, as illustrated and summarized in Fig.~\ref{fig:demo}.

Specifically, Tool-IQA adopts a systematic workflow that encompasses \textit{observation}, \textit{inspection}, and \textit{quantification}. During the initial \textit{observation} phase, a rubric note protocol is employed. The model analyzes standard imaging attributes~\cite{wang2004image}, \emph{e.g.}, sharpness, artifacts, and exposure, recording its estimated levels and textual descriptions to establish a baseline of the global visual context. Subsequently, the model is prompted to determine the sufficiency of this initial assessment. Should further scrutiny be required, a tool-assisted \textit{inspection} is initiated, selectively invoking a tool from the toolkit consisting of a Magnifier and a Gamma Corrector, to overcome the scale and contrast limitations of the initial observation. Finally, the model performs inspections upon these visual returns as auxiliary evidence for inspecting the original image and gives the assessment results.

Furthermore, to promote tool-related learning, we propose a Batch-Aware Tool-Efficiency (\textbf{BATE}) reward based on the Reinforcement Learning to Rank (RL2R) strategy \cite{wu2025visualqualityr,Guo_2025}. Different from conventional static incentives~\cite{jin2025searchr,zheng2025deepeyes}, BATE dynamically modulates rewards according to global batch-wise usage rates while enforcing intra-group competition between tool-invoked and direct assessment trajectories. This tournament-style pairwise comparison mechanism effectively guides the model to discern the need of external assistance, fostering efficient and contextual tool-calling. 

Our contributions are summarized as follows:
\begin{itemize}[itemsep=2pt, parsep=2pt, topsep=2pt, partopsep=2pt, leftmargin=1.5em]
    \item We propose Tool-IQA, which facilitates the transition from \textit{one-shot} scoring to a \textit{tool-augmented} workflow for more effective IQA with VLMs.
    \item We introduce a specialized toolkit comprising a magnifier and a Gamma corrector, designed to overcome the limitations of shallow observations by providing complementary visual evidence.
    \item We design a BATE reward function within the GRPO framework, dynamically incentivizing appropriate tool-calling to enhance the efficiency.
\end{itemize}

Extensive experiments are conducted on multiple IQA benchmarks, where Tool-IQA significantly surpasses existing methods, without the need for additional annotations or SFT starting, validating the effectiveness of our tool-augmented design. 

\section{Related Work}

\textbf{VLM-based Image Quality Assessment}. 
IQA has traditionally relied on dedicated regressors trained on Mean Opinion Scores (MOS) from subjective studies~\cite{Qin_Hu_Liu_Zheng_Liu_Li_Zhang_2023,wu2024comprehensive,zhou2025gamma}. 
Recently, VLMs have been explored as general-purpose \emph{quality judges}, capable of describing and reasoning about low-level degradations in natural language~\cite{wu2023qbench,zhang2024qbenchplus}. 
This line of work has been advanced through instruction tuning with human feedback, language-grounded descriptive or comparative assessment, distribution-aware score regression, and objectives for better cross-dataset consistency~\cite{wu2024qinstruct,you2024depicting,wu2024qalign,wu2024openended,zhu2024compare2score,you2025deqascore}. 
Further efforts incorporate region-level grounding and semantic guidance to localize distortions and assess AI-generated images with complex artifacts~\cite{chen2024qground,wang2024maagiqa}. 
Despite these advances, most existing VLM-based IQA methods remain passive and one-shot, deriving judgments from a single observation, which may miss subtle distortions or suffer from scene-level visibility and contrast biases.

\textbf{Tool-Use in VLMs}. 
Recent studies improve VLMs by augmenting them with active perception, external tools, and modular visual experts, enabling iterative evidence gathering and decision making~\cite{zheng2026deepeyes,hong2026deepeyesv2,zhang2026vipact}. 
By casting multimodal reasoning as executable programs or reasoning-and-action trajectories, models can decompose visual tasks and invoke modules, code, or external APIs for additional evidence~\cite{gupta2023visprog,suris2023vipergpt,yang2023mmreact}. 
Subsequent work strengthens this paradigm through tool-trace distillation, recursive decomposition, self-trained visual program synthesis, and robustness-oriented testing~\cite{hu2024vpd,ge2024rvp,khan2024visualreinforcement,panagopoulou2025viunit}. 
More broadly, modular tool-use and decoupled perception--reasoning frameworks show that external augmentation can complement native VLM perception, especially for tasks requiring iterative inspection, region-level evidence, or auxiliary computation~\cite{shen2023hugginggpt,qiao2024prism}. 
In this paper, we introduce tool-use into VLM-based IQA, enabling active inspection for complementary visual evidence and more calibrated quality assessment.

\section{Tool-IQA: Methodology}

\begin{figure*}[!t]
    \includegraphics[width=\linewidth]{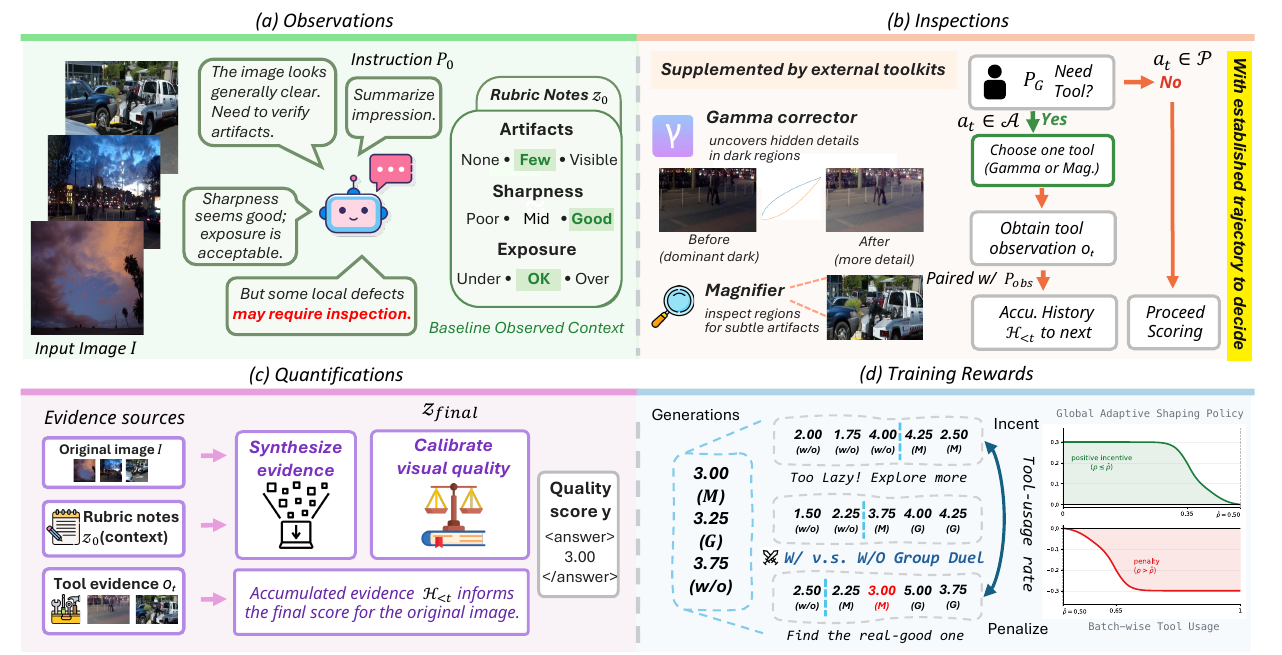}
    \caption{Overview of {Tool-IQA}, which involves specific stages, including (a) observation with rubric notes, (b) tool-augmented inspection, and (c) quantification. In addition, a (d) BATE reward is designed to improve tool-calling efficiency.}
    \label{fig:net}    
\end{figure*}

\subsection{Framework Overview}
\textbf{Problem Formulation}. We formulate Tool-IQA as a structured tool-augmented process for IQA. Different from conventional VLM-based IQA methods that rely on one-shot scoring, Tool-IQA enables the VLM to inspect quality-related cues using external tools and incorporate them as additional evidence for assessment. As illustrated in Fig.~\ref{fig:net}, given an input image $I$ and an initial task instruction $P_0$, the model first generates a rubric note $z_0$, which summarizes visual attributes, \emph{e.g.}, sharpness, exposure, contrast, and artifacts, thereby establishing the baseline. The model then performs the assessment through a trajectory $\tau = (z_0, a_1, o_1, \dots, a_T, o_T, z_{\mathrm{final}}, y)$. At each step $t$, a tool-selection prompt $P_G$ guides the model to select an operation $a_t \sim \pi_\theta(a_t \mid \mathcal{H}_{<t}, P_G)$, where $\mathcal{H}_{<t}$ denotes the accumulated history of previous assessment notes, selected operations, and observations. The operation space consists of external tool calls $\mathcal{A}$, instantiated as either the magnifier or the Gamma corrector, and a final assessment option $\mathcal{P}$. When $a_t \in \mathcal{A}$, an external tool is invoked to supplement additional visual evidence $o_t$, which is paired with an observation prompt $P_{obs}$ and provided to the VLM to support closer inspection. When $a_t \in \mathcal{P}$, the model summarizes the accumulated tool evidence together with the initial rubric note, figures the final assessment rationale $z_{\mathrm{final}}$, and outputs a calibrated quality score $y$ for the original image $I$. In this way, Tool-IQA shifts VLM-based IQA from one-shot scoring to evidence-aware quality assessment, where tool observations are explicitly incorporated to calibrate the quality score of the given image.

\textbf{Learning Objective.} We adopt GRPO \cite{Guo_2025} to train the policy $\pi_\theta$ without requiring manually annotated assessment trajectories. For a given input $(I, P_0)$, a group of $G$ trajectories $\{\tau_1, \tau_2, \dots, \tau_G\}$ is sampled from the old policy $\pi_{\theta_{\mathrm{old}}}$. The optimization objective $J(\theta)$ is formulated as:
\begin{equation}
J(\theta)=
\mathbb{E}_{\tau \sim \pi_{\theta_{\mathrm{old}}}}
\left[
\frac{\pi_\theta(\tau)}{\pi_{\theta_{\mathrm{old}}}(\tau)} A(\tau)
-\beta \cdot \mathcal{D}_{\mathrm{KL}}(\pi_\theta \parallel \pi_{\mathrm{ref}})
\right],
\end{equation}
where the relative advantage $A(\tau)$ is computed by normalizing the trajectory rewards $R(\tau)$ within each sampled group. In Tool-IQA, $R(\tau)$ combines the rank reward~\cite{wu2025visualqualityr} and the BATE reward introduced in Sec.~\ref{sec:bate}. The rank reward encourages consistency between the predicted quality score $y$ and the annotation, while the BATE reward promotes tool calls that contribute positively to the final assessment rather than simply rewarding the usage. Through these settings, GRPO assigns advantages to trajectories that achieve reliable final quantification and effective tool usage. As a result, the model is encouraged to coordinate external tools and final scoring within successful trajectories, allowing purposeful tool-calling capabilities for quality assessment to emerge through training.

\subsection{Workflow Specifics}

\textbf{Initial Rubric Observation ($z_0$)}.
As illustrated in Fig.~\ref{fig:net}(a), the assessment workflow starts with a structured decomposition of perceptual attributes. Conditioned on the initial prompt $P_{0}$, the model performs an attribute assessment over well-established quality-related dimensions~\cite{wang2004image}, \emph{i.e.}, \textbf{S}harpness, \textbf{E}xposure, \textbf{N}oise, \textbf{A}rtifacts, and \textbf{V}isibility, and records the result as a structured rubric note $z_0$. Rather than directly producing a score, $z_0$ contains categorical attribute levels and brief observed notes, following the format:
\begin{center}
\texttt{Attribute: [Level] - Brief Note}.
\end{center}
This rubric analysis establishes a perceptual baseline for the subsequent assessment by eliciting and organizing the rationales into structured notes, as shown in Fig.~\ref{fig:net}. These notes summarize and open up quality-related entries to be enriched by external tool observations, thereby benefiting final score calibration. The prompt template is detailed in \textbf{Appendix~\ref{sec:app_prompt}}.

\textbf{Tool-Augmented Inspection ($a_t$)}.
Following the rubric analysis, Tool-IQA introduces a decision step in which the model chooses whether to acquire additional tool-assisted observations before final scoring, as illustrated in Fig.~\ref{fig:net}(b). Guided by the tool-selection prompt $P_G$, the model is instructed to choose one of the following operations:
\begin{center}
\texttt{Output one of:}\
\texttt{<tool\_call>} or \texttt{<answer>}.
\end{center}
Under such, the model may select a tool call ($a_t \in \mathcal{A}$) to perform further inspections, or select the quality assessment option ($a_t \in \mathcal{P}$) to provide the final calibrated score. Generated by the policy $\pi_\theta$ conditioned on the accumulated history, this decision determines how the assessment moves from inspection to quantification. In this way, Tool-IQA enables purposeful tool-calling behaviors to emerge during training, allowing the assessment process to adapt to the available visual evidence.

\textbf{Quantification ($z_{\mathrm{final}}, y$)}.
When the policy emits the quality assessment option ($a_t \in \mathcal{P}$), or when the maximum interaction depth is reached, the model directly provides the final calibrated score, as illustrated in Fig.~\ref{fig:net}(c). Specifically, the model synthesizes the accumulated assessment trajectory by combining the initial rubric note $z_0$ with any supplementary tool observations $\{o_t\}$ gathered during the inspection. It then generates the final assessment rationale $z_{\mathrm{final}}$ and outputs a scalar quality score $y$ for the original image $I$. In this way, the final prediction is calibrated with both the initial assessment baseline and external tool observations.

\subsection{Perceptual Toolkit}
To support observation supplementation, we introduce two optional external tools to provide complementary views: the \textit{Magnifier} enables scale-aware inspection by re-concentrating representation on regions, while the \textit{Gamma Corrector} provides visibility views. 

\textbf{Magnifier (Scale-aware Inspection).}
In the original view, local distortions may be under-emphasized due to inherently limited spatial extent or interference from contents, \emph{e.g.}, subtle blur, texture degradation, blocking artifacts, and small structural distortions. To supplement the observations, the model issues a magnifier call with a candidate region specified by a bounding box $[x_1, y_1, x_2, y_2]$, conditioned on the accumulated assessment context. The corresponding region is cropped from the original image and presented as an additional view for inspection. Such a view changes the effective inspection scale, making regional structures, textures, and artifacts more salient to the VLM. The resulting observation provides additional evidence supporting the final score calibration.

\textbf{Gamma Corrector (Visibility Observation).}
The Gamma Corrector is also available as an optional external tool for generating visibility-referenced views under different intensity conditions. When selected by the model, it supports the inspection of content visibility in low-contrast regions, especially for extremely dark or bright images, where details may be difficult to assess from the original view. To obtain such observations, the Gamma Corrector applies a non-linear intensity transformation, denoted as $I_{\mathrm{out}} = I_{\mathrm{in}}^{\gamma}$, where $I_{\mathrm{in}}$ is normalized to $[0,1]$ and $\gamma$ controls the intensity remapping. By varying $\gamma$, the tool produces alternative views that emphasize different intensity ranges, with generated views treated as supplementary visual references that help the model assess visibility-referenced cues and calibrate the final quality score of the original image.

\subsection{Batch-Aware Tool-Efficiency Reward}
\label{sec:bate}
To encourage purposeful tool calling, we propose a \textbf{B}atch-\textbf{A}ware \textbf{T}ool-\textbf{E}fficiency Reward, denoted as \textbf{BATE}. Instead of treating tool usage as a binary objective, BATE evaluates the \textit{\textbf{marginal utility}} of tool calls relative to tool-free scoring, encouraging external tools to be invoked only when they provide positive contributions to the final assessment. As shown in Fig.~\ref{fig:net}(d), BATE operates at two levels: a local (\emph{i.e.}, within the generation group) pairwise comparison of solution fidelity, and a global (\emph{i.e.}, across all groups) batch-level tool-calling rate regulation.

\textbf{Tool-Free Duels}. 
To promote efficiency and discourage indiscriminate tool calling, BATE evaluates tool usage by its \textit{\textbf{gain}} over tool-free scoring. Rather than assigning a bonus to any tool call, the generated trajectories for each input are partitioned into two patterns: a tool-invoked set $\mathcal{T}$ and a tool-free set $\mathcal{N}$. The utility of tool usage is then estimated through pairwise duels between these two cohorts, following a Thurstonian preference formulation. 

Specifically, for a tool-invoked trajectory $t \in \mathcal{T}$, the base reward is derived by comparing it against the local ensemble of tool-free trajectories $\mathcal{N}$ generated for the same input:
\begin{equation}
R_{\text{base}}^{(t)} \propto \frac{1}{|\mathcal{N}|} \sum_{n \in \mathcal{N}} 
\sqrt{\Phi\left(\frac{s_t - s_n}{\tau}\right)},
\end{equation}
where $\Phi$ denotes the cumulative distribution function of the normal distribution, $\tau$ is a temperature parameter, and $s_t$ and $s_n$ denote the fidelity rank scores of the tool-invoked and tool-free trajectories, respectively, computed following~\cite{wu2025visualqualityr}. By grounding the reward in such group-wise duels, BATE reinforces tool usage only when it brings a positive accuracy margin over tool-free scoring.

\textbf{Global Adaptive Shaping}.
During training, the model may fall into degenerate patterns, \emph{e.g.}, indiscriminate tool calling, leading to insufficient exploration or unnecessary computation and inspection on high-quality images with few visible degradations. To address this, we compute the global tool utilization rate $\rho$ over all valid samples in the current batch and use it to modulate the reward through two schedules: an \textit{incentive curve} and a \textit{penalty curve}. When tool utilization is sparse ($\rho \leq \hat{\rho}$), the incentive term encourages tool exploration. Conversely, when the utilization is saturated ($\rho > \hat{\rho}$), the penalty term makes tool calling more selective by suppressing low-margin usage. This encourages tool-invoked trajectories to retain a sufficient advantage over their tool-free counterparts. The total reward is defined as:
\begin{equation}
R_{\text{total}}^{(t)} = R_{\text{base}}^{(t)} +
\begin{cases}
\mathcal{S}_{\text{inc}}(\rho, s_t) & \text{if } \rho \leq \hat{\rho}, \\
-\mathcal{S}_{\text{pen}}(\rho) & \text{if } \rho > \hat{\rho},
\end{cases}
\end{equation}
where $\mathcal{S}_{\text{inc}}$ and $\mathcal{S}_{\text{pen}}$ denote smoothed sigmoidal piecewise functions for incentive and penalty, respectively. By batch-level regulation, BATE encourages a balanced and purposeful tool-calling policy, where external tools are favored when they provide a factual contribution to the final assessment.

\section{Experiments}
\subsection{Experimental Settings}

\textbf{Experiment Setup}. Following previous protocols~\cite{li2025qinsight,wu2025visualqualityr}, we train Tool-IQA on $60\%$ of the synthetic KADID-10K dataset~\cite{lin2019kadid}, which contains 10,125 images with 25 distortion types. Without further fine-tuning, we evaluate it on four authentic blind IQA benchmarks, including SPAQ~\cite{SPAQ}, CLIVE~\cite{LIVEC}, BID~\cite{BID}, and KonIQ-10K~\cite{hosu2020koniq}, as well as three algorithm-oriented datasets, Dehaze~\cite{8654007}, SRIQA~\cite{11094894}, and Deblur~\cite{10.1145/2508363.2508391}. These datasets cover diverse real-world distortions, capture conditions, and task-specific degradations, enabling a comprehensive evaluation of Tool-IQA.

\textbf{Evaluation Metrics}. We evaluate performance using Spearman's rank-order correlation coefficient (SRCC) and Pearson's linear correlation coefficient (PLCC). SRCC measures monotonic consistency with human judgments, while PLCC measures linear correlation after score regression. For both metrics, higher values indicate better performance and alignment with human annotations.

\begin{table*}[!t]
\centering
\caption{Results of models trained on KADID-10K. Gains are against the best competing method. VisualQuality-R1 is reproduced with Qwen3. The top twos are in \textbf{bold} and \underline{underline}, respectively.}
\resizebox{\textwidth}{!}{
\begin{tabular}{lcccccccc}
\toprule
\multicolumn{1}{c}{\multirow{2}{*}{\textbf{Methods}}} 
& \multicolumn{8}{c}{\textbf{\textit{Scenarios}} - \textbf{SRCC/PLCC}} \\
\cmidrule(lr){2-9}
 & BID & CLIVE & KonIQ & SPAQ & \shortstack{Deblur} & \shortstack{SRIQA} & \shortstack{Dehaze} & Avg \\ 
\midrule
\multicolumn{9}{l}{\cellcolor[HTML]{EFEFEF}\textit{Handcrafted}} \\
NIQE~\cite{mittal2012making}     
& .515/.527 & .450/.494 & .421/.439 & .676/.683 & .360/.376 & .557/.587 & .343/.482 & .475/.513 \\
BRISQUE~\cite{mittal2012no}      
& .522/.528 & .314/.362 & .385/.400 & .614/.624 & .389/.380 & .482/.556 & .242/.468 & .421/.474 \\ 
\midrule
\multicolumn{9}{l}{\cellcolor[HTML]{EFEFEF}\textit{Discriminative Deep-Learning-based}} \\
UNIQUE~\cite{zhang2021uncertainty} 
& .412/.385 & .470/.472 & .649/.590 & .751/.708 & .669/.654 & .649/.668 & .577/.578 & .597/.579 \\
MUSIQ~\cite{ke2021musiq}           
& .327/.280 & .284/.325 & .473/.435 & .720/.666 & .656/.563 & .404/.441 & .458/.455 & .475/.452 \\
MANIQA~\cite{yang2022maniqa}       
& .420/.512 & .487/.571 & .213/.257 & .745/.753 & .726/.728 & .263/.243 & .608/.663 & .495/.532 \\
DEIQT~\cite{Qin_Hu_Liu_Zheng_Liu_Li_Zhang_2023} 
& .534/.539 & .511/.528 & .498/.493 & .740/.735 & .734/.759 & .412/.438 & .613/.652 & .577/.592 \\ 
\midrule
\multicolumn{9}{l}{\cellcolor[HTML]{EFEFEF}\textit{VLM-based}} \\
LIQE~\cite{zhang2023blind}         
& .677/.680 & .719/.726 & .684/.652 & .815/.814 & .797/.712 & \underline{.743}/.775 & .646/.661 & .726/.717 \\
Q-Align~\cite{wu2024qalign}        
& .576/.651 & .554/.643 & .573/.612 & .767/.779 & .761/.802 & .684/.713 & .455/.525 & .624/.675 \\
DeQA-Score~\cite{you2025deqascore} 
& .702/.743 & .743/.795 & .677/.703 & .852/.858 & .785/.838 & .710/.763 & .643/.688 & .730/.770 \\
Qwen2.5-VL-7B~\cite{qwen2025qwen25technicalreport} 
& .711/.725 & .733/.760 & .754/.810 & .848/.854 & .820/.852 & .603/.653 & .458/.553 & .704/.744 \\
Qwen3-VL-8B-Ins~\cite{yang2025qwen3technicalreport} 
& .779/.725 & .760/.772 & .729/.769 & .856/.849 & .810/.842 & .694/.797 & .603/.652 & .747/.772 \\
Q-Insight~\cite{li2025qinsight} 
& .784/.796 & .761/.795 & \underline{.806/.829} & \underline{.872/.872} & \underline{.831}/.857 & .724/.798 & .601/.669 & .768/.802 \\
VisualQuality-R1~\cite{wu2025visualqualityr}        
& \underline{.788/.807} & \underline{.792/.799} & .788/.811 & \underline{.872}/.856 & .811/\underline{.859} & .709/\underline{.814} & \underline{.662/.721} & \underline{.775/.810} \\
\midrule
\rowcolor[HTML]{E8F4FD} \textit{Tool-IQA (Ours)} 
& \textbf{.828/.842} & \textbf{.833/.854} & \textbf{.825/.848} & \textbf{.898/.898} & \textbf{.848/.878} & \textbf{.751/.847} & \textbf{.730/.797} & \textbf{.816/.852} \\
\rowcolor[HTML]{F2F8FD} \textit{Gains} 
& $.040/.035$ & $.041/.055$ & $.019/.019$ & $.026/.026$ & $.017/.019$ & $.008/.033$ & $.068/.076$ & $.042/.042$ \\
\bottomrule
\end{tabular}
}
\label{tab:cross-datasets}
\end{table*}

\begin{table*}[!t]
\centering

\begin{minipage}[t]{0.43\textwidth}
\vspace{0pt}
\centering
\caption{Replacement study on VQ-R1.}
\label{tab:replacement_study}
\resizebox{\linewidth}{!}{
\begin{tabular}{lccc}
\toprule
\multicolumn{1}{c}{\textbf{Method}}
& \multicolumn{1}{c}{\textbf{CLIVE}}
& \multicolumn{1}{c}{\textbf{KonIQ}}
& \multicolumn{1}{c}{\textbf{SPAQ}} \\
\midrule
VQ-R1
& .792/.799 & .788/.811 & .872/.856 \\
\# Replaced
& 476 & 2357 & 6243 \\
VQ-R1 w/ Tools
& .794/.807 & .784/.808 & .870/.858 \\
\midrule
\rowcolor[HTML]{E8F4FD}
\textit{Tool-IQA (Ours)}
& \textbf{.833/.854} & \textbf{.825/.848} & \textbf{.898/.898} \\
\rowcolor[HTML]{F2F8FD}
\textit{Gains} 
& $.039/.047$ & $.041/.040$ & $.028/.040$ \\
\bottomrule
\end{tabular}
}
\end{minipage}
\hfill
\begin{minipage}[t]{0.55\textwidth}
\vspace{0pt}
\centering
\caption{Augmented methods trained on KonIQ.}
\label{tab:tool-iqa}
\resizebox{\linewidth}{!}{
\begin{tabular}{lccccc}
\toprule
\multicolumn{1}{c}{\multirow{1}{*}{\textbf{Method}}}
& \multicolumn{1}{c}{\textbf{SPAQ}} 
& \multicolumn{1}{c}{\textbf{KADID}}
& \multicolumn{1}{c}{\textbf{CLIVE}} 
& \multicolumn{1}{c}{\textbf{PIPAL}~\cite{10.1007/978-3-030-58621-8_37}} \\
\midrule
VisualQuality-R1
& .892/.889 & \underline{.712/.703} & .827/.856 & .441/.451 \\
\multicolumn{5}{l}{\cellcolor[HTML]{EFEFEF}\textit{Tool-augmented}} \\
Zoom-IQA~\cite{liang2026zoomi}
& \underline{.900/.902} & .700/.701 & \underline{.870/.887} & \underline{.465/.468} \\
\midrule
\rowcolor[HTML]{E8F4FD}
\textit{Tool-IQA (Ours)} 
& \textbf{.905/.906} & \textbf{.721/.737} & \textbf{.889/.902} & \textbf{.479/.483} \\
\rowcolor[HTML]{F2F8FD}
\textit{Gains} 
& $.005/.004$ & $.021/.036$  & $.019/.015$  & $.014/.015$\\
\bottomrule
\end{tabular}
}
\end{minipage}

\end{table*}

\textbf{Implementation Details}.
{\textit{{Qwen3-VL-8B-Instruct}}} was used as the foundation VLM. 
Training was conducted with GRPO for 680 steps on 16 NVIDIA A800 GPUs, where 4 GPUs were used for vLLM-based on-policy sampling and 12 for parameter updates, taking about 16 hours. During the first 340 steps, we used a learning rate of $2 \times 10^{-6}$, KL coefficient $\beta=0.04$, temperature $0.9$, 4 candidates per input, warmup ratio $0.05$, batch size $12$, and sequence-level importance sampling. In the last 340 steps, only $\beta$ was increased to $0.06$. The model was allowed at most two tool-use rounds, with one call per round. Magnifier coordinates were parsed from \texttt{<tool\_call>}, while the Gamma Corrector provided darkening and brightening options. Invalid tool calls were replaced with failure prompts, and the random seed was set to 42.

The training objective is a combination of BATE, the rank reward~\cite{wu2025visualqualityr}, and a format constraint. For BATE, the target tool utilization rate was $\hat{\rho}=0.50$. Batch-level utilization $\rho$ was shaped with incentives for $\rho \in [0.35,0.50)$ and penalties for $\rho \in (0.50,0.65]$, with threefold strength when $\rho$ fell outside $[0.35,0.65]$. The shaping signals were sigmoid-smoothed, rank-modulated, and applied only to successful tool calls.

\begin{table*}[!t]
\centering
\setlength{\tabcolsep}{3pt}
\renewcommand{\arraystretch}{0.95}

\begin{minipage}[t]{0.48\textwidth}
\centering
\caption{Ablation study on the tool-calling pattern regarding training.}
\label{tab:ablation-training-updated}
\resizebox{\linewidth}{!}{
\begin{tabular}{lcccc}
\toprule
\multicolumn{1}{l}{\multirow{2}{*}{\textbf{Methods}}} & 
\multirow{2}{*}{\textbf{Training}} & 
\multicolumn{3}{c}{\textbf{\textit{Scenarios}}} \\
\cmidrule(lr){3-5}
 & & CLIVE & KonIQ & SPAQ \\ 
\midrule
\multicolumn{5}{l}{\textbf{SRCC}} \\
Qwen3-VL-8B-Ins & $\times$ & 0.760 & 0.729 & 0.856 \\
VisualQuality-R1 & $\checkmark$ & 0.792 & \underline{0.788} & \underline{0.872} \\
Tool-IQA & $\times$ & \underline{0.794} & 0.732 & 0.866 \\
\rowcolor[HTML]{E8F4FD}Tool-IQA & $\checkmark$ & \textbf{0.833} & \textbf{0.825} & \textbf{0.898} \\
\midrule
\multicolumn{5}{l}{\textbf{PLCC}} \\
Qwen3-VL-8B-Ins & $\times$ & 0.772 & 0.769 & 0.849 \\
VisualQuality-R1 & $\checkmark$ & 0.799 & \underline{0.811} & 0.856 \\
Tool-IQA & $\times$ & \underline{0.828} & 0.781 & \underline{0.872} \\
\rowcolor[HTML]{E8F4FD}Tool-IQA & $\checkmark$ & \textbf{0.854} & \textbf{0.848} & \textbf{0.898} \\
\bottomrule
\end{tabular}
}
\end{minipage}
\hfill
\begin{minipage}[t]{0.50\textwidth}
\centering
\caption{Ablation study on the effectiveness of tools. \textbf{Mag.} and \textbf{Cor.} denote the Magnifier and Gamma Corrector, respectively.}
\label{tab:ablation-tools-modified}
\resizebox{\linewidth}{!}{
\begin{tabular}{lccccc}
\toprule
\multicolumn{1}{l}{\multirow{2}{*}{\textbf{Methods}}} & 
\multirow{2}{*}{\textbf{Mag.}} & 
\multirow{2}{*}{\textbf{Cor.}} & 
\multicolumn{3}{c}{\textbf{\textit{Scenarios}}} \\
\cmidrule(lr){4-6}
 & & & CLIVE & KonIQ & SPAQ \\ 
\midrule
\multicolumn{6}{l}{\textbf{SRCC}} \\
VisualQuality-R1 & $\times$ & $\times$ & 0.792 & \underline{0.788} & 0.872 \\
Tool-IQA & $\checkmark$ & $\times$ & \underline{0.808} & 0.779 & \underline{0.880} \\
\rowcolor[HTML]{E8F4FD}Tool-IQA & $\checkmark$ & $\checkmark$ & \textbf{0.833} & \textbf{0.825} & \textbf{0.898} \\
\midrule
\multicolumn{6}{l}{\textbf{PLCC}} \\
VisualQuality-R1 & $\times$ & $\times$ & 0.799 & \underline{0.811} & 0.856 \\
Tool-IQA & $\checkmark$ & $\times$ & \underline{0.833} & 0.809 & \underline{0.882} \\
\rowcolor[HTML]{E8F4FD}Tool-IQA & $\checkmark$ & $\checkmark$ & \textbf{0.854} & \textbf{0.848} & \textbf{0.898} \\
\bottomrule
\end{tabular}
}
\end{minipage}
\end{table*}

\subsection{Performance Comparison}

\textbf{Quantitative Comparison}. As shown in Tab.~\ref{tab:cross-datasets}, we compare Tool-IQA with representative handcrafted~\cite{mittal2012making,mittal2012no}, discriminative learning-based~\cite{zhang2021uncertainty, ke2021musiq, yang2022maniqa, Qin_Hu_Liu_Zheng_Liu_Li_Zhang_2023}, and VLM-based methods~\cite{zhang2023blind, wu2024qalign, you2025deqascore, yang2025qwen3technicalreport, wu2025visualqualityr, li2025qinsight}. We see that conventional models show limited cross-dataset generalization, whereas VLM-based methods achieve much more competitive results. GRPO-based methods generally outperform SFT-based ones, suggesting the effectiveness of RLHF for VLM-based IQA. Among them, Tool-IQA consistently obtains superior performance, especially on authentic datasets with complex contents and capture conditions, such as CLIVE and SPAQ. In particular, Tool-IQA achieves a PLCC of 0.854 on CLIVE. The gains mainly come from the tool-augmented workflow, which provides supplementary visual observations for better score calibration. To ensure a fair comparison, we further equip VisualQuality-R1 (VQ-R1) \cite{wu2025visualqualityr} with the tool outputs of Tool-IQA and let it analyze the replaced samples. The results are shown in Tab.~\ref{tab:replacement_study}. We see that VQ-R1 w/ Tools brings only marginal gains, indicating limited ability to leverage external tools. In addition, we present more results of Tool-IQA on LIVEFB and TID datasets in \textbf{Appendix~B}.

\textbf{Comparison with Tool-augmented Methods}.
We further compare Tool-IQA with a recently developed tool-augmented method, Zoom-IQA~\cite{liang2026zoomi}. Since the source code and model of Zoom-IQA are not publicly available, we re-train our Tool-IQA on KonIQ according to the experimental settings of Zoom-IQA, and compare with its officially reported results. As shown in Tab.~\ref{tab:tool-iqa}, Tool-IQA achieves consistently better performance without requiring additional cold-starting annotations, demonstrating the effectiveness of our design, including the toolkits and BATE reward.

\begin{figure*}[!t]
    \centering
    \includegraphics[width=1\linewidth]{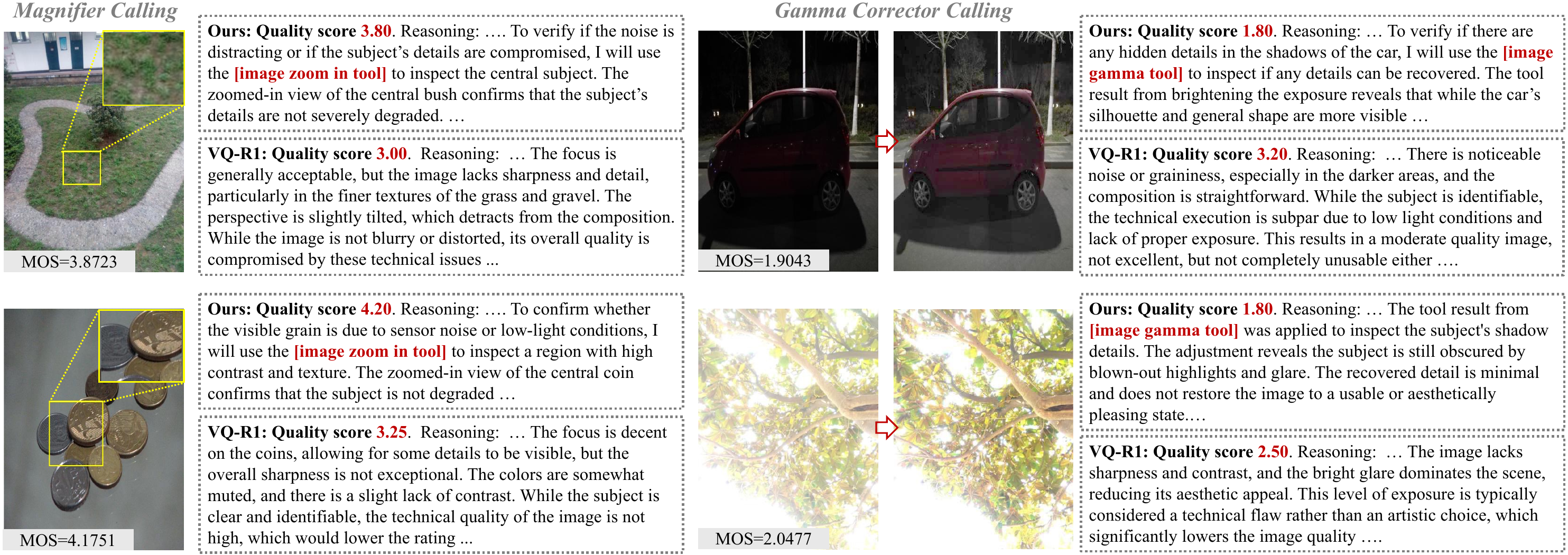}
    \caption{Visualization of the tool-calling mechanism of Tool-IQA. Tool-IQA demonstrates higher score prediction consistency with the ground-truths while providing better assessment process.}
    \label{fig:qualitative_results}
    \vspace{-2mm}
\end{figure*}

\textbf{Visualizations and Qualitative Analysis}.
As illustrated in Fig.~\ref{fig:qualitative_results}, Tool-IQA learns to invoke tools within the assessment trajectory according to the input image. The Magnifier is typically used to inspect details, \emph{e.g.}, to examine textures, or artifacts. In contrast, the Gamma Corrector provides visibility-aware views for images with ill-exposed, or low-contrast regions, specifically, to make structures and visibility-related cues more discernible for final score calibration. This behavior resembles human visual inspection, where observers adjust their viewing focus or sensitivity according to image content. Compared with VQ-R1, Tool-IQA produces predictions that are more consistent with human annotations, which is also reflected in the quantitative results.

\subsection{Ablation Studies}

\textbf{Tool-calling Patterns}.
The proposed tool-augmented workflow and the training pipeline are further examined in Tab.~\ref{tab:ablation-training-updated}. Even without training, Tool-IQA achieves better correlations than the baseline, suggesting that the structured workflow can elicit useful tool-assisted observations from the VLM. Further gains are obtained after training with the proposed reward functions, consistently outperforming VQ-R1 across all scenarios, and confirming that robust quality assessment benefits from the combination of augmented observation supplementation and purposeful tool-calling optimization.

\begin{table*}[!t]
\centering
\setlength{\tabcolsep}{3pt}
\renewcommand{\arraystretch}{0.95}

\begin{minipage}[t]{0.49\textwidth}
\centering
\caption{Comparative analysis of VQ-R1 and Tool-IQA on SPAQ \textbf{w/ training}.}
\label{tab:mechanism-spaq-trained}
\resizebox{\linewidth}{!}{
\begin{tabular}{cc|cccc|cc}
\toprule
\multicolumn{2}{c|}{\textbf{VQ-R1}} &
\multicolumn{4}{c|}{\textbf{Tool-IQA}} &
\multicolumn{2}{c}{\textbf{Gain ($\Delta$)}} \\
\cmidrule(lr){1-2} \cmidrule(lr){3-6} \cmidrule(lr){7-8}
\small SRCC & \small PLCC &
\small $\mathcal{F}$(\%) & $\mathcal{T}$ & \small SRCC & \small PLCC &
\small SRCC & \small PLCC \\
\midrule

\multicolumn{8}{c}{\textit{Mechanism Analysis}} \\
0.872 & 0.856 & 100.0 & --- & \textbf{0.898} & \textbf{0.898} & +0.026 & +0.042 \\
0.501 & 0.492 & 43.9 & \textcolor{red}{$\times$} & \textbf{0.586} & \textbf{0.574} & +0.085 & +0.083 \\
0.752 & 0.723 & 56.1 & \textcolor{blue}{$\checkmark$} & \textbf{0.790} & \textbf{0.777} & +0.039 & +0.054 \\
\midrule

\multicolumn{8}{c}{\textit{Specific Tool Effectiveness}} \\
0.762 & 0.731 & 47.4 & $\mathcal{C}$ & \textbf{0.800} & \textbf{0.782} & +0.038 & +0.051 \\
0.677 & 0.678 & 8.7 & $\mathcal{M}$ & \textbf{0.695} & \textbf{0.740} & +0.018 & +0.063 \\
\bottomrule
\end{tabular}
}
\end{minipage}
\hfill
\begin{minipage}[t]{0.49\textwidth}
\centering
\caption{Analysis of tool-use capability of Qwen3-VL and Tool-IQA on SPAQ \textbf{w/o training}.}
\label{tab:zeroshot-spaq}
\resizebox{\linewidth}{!}{
\begin{tabular}{cc|cccc|cc}
\toprule
\multicolumn{2}{c|}{\textbf{Qwen3-VL}} &
\multicolumn{4}{c|}{\textbf{Tool-IQA (w/o Train)}} &
\multicolumn{2}{c}{\textbf{Gain ($\Delta$)}} \\
\cmidrule(lr){1-2} \cmidrule(lr){3-6} \cmidrule(lr){7-8}
\small SRCC & \small PLCC &
\small $\mathcal{F}$(\%) & $\mathcal{T}$ & \small SRCC & \small PLCC &
\small SRCC & \small PLCC \\
\midrule

\multicolumn{8}{c}{\textit{Mechanism Analysis}} \\
0.857 & 0.850 & 100.0 & --- & \textbf{0.866} & \textbf{0.872} & +0.010 & +0.022 \\
\textbf{0.249} & 0.395 & 11.1 & \textcolor{red}{$\times$} & 0.172 & \textbf{0.423} & -0.076 & +0.028 \\
0.840 & 0.825 & 88.9 & \textcolor{blue}{$\checkmark$} & \textbf{0.849} & \textbf{0.852} & +0.009 & +0.026 \\
\midrule

\multicolumn{8}{c}{\textit{Specific Tool Effectiveness}} \\
0.843 & 0.827 & 75.7 & $\mathcal{C}$ & \textbf{0.847} & \textbf{0.854} & +0.004 & +0.028 \\
0.600 & 0.689 & 13.5 & $\mathcal{M}$ & \textbf{0.603} & \textbf{0.715} & +0.003 & +0.027 \\
\bottomrule
\end{tabular}
}
\end{minipage}
\end{table*}

\textbf{Tool Effectiveness}.
The effectiveness of the proposed tools, \emph{i.e.}, the Magnifier and Gamma Corrector, is validated in Tab.~\ref{tab:ablation-tools-modified}. The integration of such tools consistently improves performance over the baseline after training, with the best results achieved when both tools are available. Rather than relying solely on the original image view, the Magnifier and Gamma Corrector provide complementary auxiliary views for scale-aware inspection and visibility-aware observation, respectively. These observations complement the initial analysis and contribute to more reliable final score calibration. Results of Qwen3 under zero-shot settings could be found in \textbf{Appendix~C}.

\begin{figure*}[!t]
\centering
\setlength{\tabcolsep}{3pt}
\renewcommand{\arraystretch}{0.95}

\begin{minipage}[t]{0.45\textwidth}
\centering

\captionof{table}{Efficiency comparison.}
\label{tab:efficiency}
\resizebox{0.95\linewidth}{!}{
\begin{tabular}{lccc}
\toprule
\textbf{Method} & \textbf{Speed (it/s)} & \textbf{OTokens} & \textbf{Tool Calls} \\
\midrule
VQ-R1 & 6.35 & 192.85 & - \\
\midrule
w/o BATE & 2.28 & 587.62 & 1.77 \\
\rowcolor[HTML]{E8F4FD}
Tool-IQA & 4.10 & 350.51 & 0.39 \\
\bottomrule
\end{tabular}
}

\captionof{table}{Ablation study on components.}
\label{tab:ablation-components}
\resizebox{\linewidth}{!}{
\begin{tabular}{ccc|cc|cc}
\toprule
\multicolumn{3}{c|}{\textbf{\textit{Components}}} & 
\multicolumn{2}{c|}{\textbf{CLIVE}} & 
\multicolumn{2}{c}{\textbf{SPAQ}} \\
\cmidrule(lr){1-3} \cmidrule(lr){4-5} \cmidrule(lr){6-7}
Rubric & Tools & BATE & SRCC & PLCC & SRCC & PLCC \\ 
\midrule
$\times$ & $\times$ & $\times$ & 0.792 & 0.799 & 0.872 & 0.856 \\
$\checkmark$ & $\times$ & $\times$ & 0.796 & 0.814 & 0.878 & 0.873 \\
$\checkmark$ & $\checkmark$ & $\times$ & \underline{0.812} & \underline{0.837} & \underline{0.888} & \underline{0.890} \\
\rowcolor[HTML]{E8F4FD} 
$\checkmark$ & $\checkmark$ & $\checkmark$ & \textbf{0.832} & \textbf{0.854} & \textbf{0.898} & \textbf{0.898} \\
\bottomrule
\end{tabular}
}

\end{minipage}
\hfill
\begin{minipage}[t]{0.52\textwidth}
\centering
\vspace{0pt}
\includegraphics[width=\linewidth]{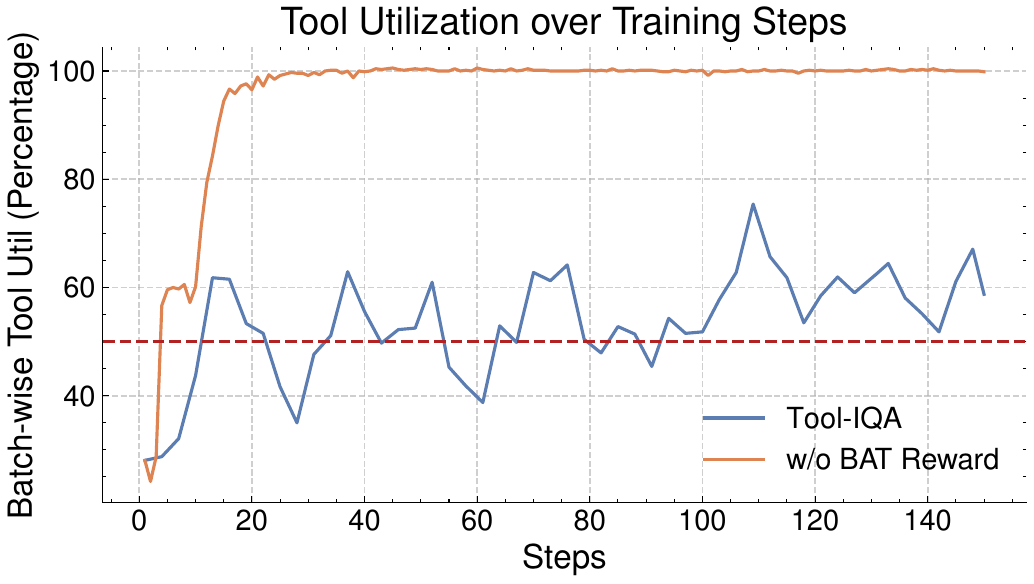}
\vspace{-4mm}
\captionof{figure}{Curves of batch-wise tool utilization over training steps.}
\label{fig:wobat}
\end{minipage}

\end{figure*}

\textbf{Tool Utilization \& Efficiency}.
Tab.~\ref{tab:mechanism-spaq-trained} summarizes tool-use behavior on SPAQ, where $\mathcal{F}$ denotes frequency, $\mathcal{T}$ tool usage, and $\mathcal{C}$/$\mathcal{M}$ Gamma Corrector/Magnifier. With BATE, Tool-IQA learns selective tool invocation, reducing unnecessary computation while improving over VQ-R1 when tools are used. The Gamma Corrector yields the largest gain, with a $\Delta$ of +0.038, while gains on tool-free samples ($\mathcal{T}=\times$) indicate that the rubric-note assessment and BATE learning also benefit the tool-free pathway. Tab.~\ref{tab:zeroshot-spaq} shows that the base VLM tends to call tools before training, with a calling rate of 89\%. This indicates basic instruction-following ability for tool use, but also reveals poor IQA-specific calibration, calling for training and the BATE promotion. Furthermore, we evaluate the inference latency of full pipeline, including tool generations and feeding, as shown in Tab.~\ref{tab:efficiency}. Tool-IQA further improves speed and reduces output tokens and tool calls compared with w/o BATE, demonstrating better efficiency and supporting the component gains in Tab.~\ref{tab:ablation-components}.

\textbf{Modular Study}.
Tab.~\ref{tab:ablation-components} dissects the contribution of each core component, showing consistent gains from the rubric-note protocol, tool-augmented workflow, and BATE reward. As shown in Fig.~\ref{fig:wobat}, removing BATE causes excessive tool calling, while BATE promotes selective tool use by favoring calls with positive marginal contributions.

\section{Conclusion}
In this work, we presented Tool-IQA, which reformulated VLM-based IQA from \textit{one-shot} prediction into a \textit{tool-augmented} assessment workflow. Specifically, we introduced two external tools, including a Magnifier for scale-aware inspection and a Gamma Corrector for visibility-aware observation, to supplement with complementary visual evidence. In this way, Tool-IQA produced calibrated and better-aligned quality scores for the input image. Furthermore, with the proposed BATE reward under GRPO, Tool-IQA learned purposeful tool-calling behaviors via principled exploration. Extensive experiments on widely used IQA benchmarks demonstrated that Tool-IQA significantly outperformed other VLM-based IQA models, without requiring additional datasets or annotation reconstruction. 

\textbf{Limitations}. Despite promising results, our work faces some constraints. First, the framework relies on the \textit{capabilities} of the base VLM. Second, training efficiency is limited by the \textit{rollout sampling speed}. To push the boundary of VLM-based IQA, foundation development needs to be pursued.

{\small
\bibliographystyle{plain}
\bibliography{neurips_2026}
}

\clearpage
\appendix

\section*{Appendix}

In this appendix, we provide the following materials:
\begin{itemize}
    \item[\textbf{A}] Details of prompts used in Tool-IQA, including those providing tools to VLMs.
    \item[\textbf{B}] More results of Tool-IQA on LIVEFB and TID datasets.
    \item[\textbf{C}] More studies on the effectiveness of tool callings and relations to rubric notes.
\end{itemize}

\section{Prompt Listings}
\textbf{Rubric Observation}.
\label{sec:app_prompt}
We use the following prompts for the rubric turn that conditions the model.

\begin{tcolorbox}[
  title=Rubric Reasoning Prompt,
  colback=gray!3,
  colframe=gray!50,
  boxrule=0.6pt,
  arc=2mm,
  left=1.5mm,right=1.5mm,top=1mm,bottom=1mm,
  fonttitle=\bfseries,
  enhanced,
  breakable
]
\begin{lstlisting}[style=promptstyle]
You are doing the image quality assessment task.
Instead of giving a score immediately, first analyze the image based on the following key dimensions:

1. **Sharpness & Detail**: Is the image blurry or sharp? Are textures well-defined?
2. **Noise & Grain**: Is there visible noise (luminance/color)? Is it distracting?
3. **Artifacts**: Are there compression artifacts (JPEG blocks), ringing, or ghosting?
4. **Exposure & Color**: Is it over/underexposed? Are colors natural?
5. **Saliency Visibility**: Is the MAIN SUBJECT clearly visible? Or is it obscured by shadows/highlights? Focus primarily on the subject's clarity, treating background visibility as secondary.

Output Format:
<thinking>
...
</thinking>
<answer>
Sharpness: [Low/Medium/High] - Brief note
Noise: [None/Visible/Severe] - Brief note
Artifacts: [None/Minor/Severe] - Brief note
Exposure: [Poor/Good/Excellent] - Brief note
Saliency Visibility: [Subject Clear/Obscured] -  Brief Note
</answer>
\end{lstlisting}
\end{tcolorbox}

\textbf{Action Gated - Tool or Answer}.
We use the following prompt template to choose actions.

\begin{tcolorbox}[
  title=Rubric Reasoning Prompt,
  colback=gray!3,
  colframe=gray!50,
  boxrule=0.6pt,
  arc=2mm,
  left=1.5mm,right=1.5mm,top=1mm,bottom=1mm,
  fonttitle=\bfseries,
  enhanced,
  breakable
]
\begin{lstlisting}[style=promptstyle]
Single-Action per Turn: choose EXACTLY ONE
- A) Final answer (1.00 - 5.00)
- B) Tool call

Output one of:

A) Final Answer:
<thinking>...</thinking>
<answer>...</answer>

B) Tool call:
<thinking>...</thinking>
<tool_call>...</tool_call>
\end{lstlisting}
\end{tcolorbox}

\textbf{Providing Tools}.
Following are the prompt template for tool providing, as per Qwen3~\cite{yang2025qwen3technicalreport}. Models are required to answer with JSON to call tools.

\begin{tcolorbox}[
  title=Rubric Reasoning Prompt,
  colback=gray!3,
  colframe=gray!50,
  boxrule=0.6pt,
  arc=2mm,
  left=1.5mm,right=1.5mm,top=1mm,bottom=1mm,
  fonttitle=\bfseries,
  enhanced,
  breakable
]
\begin{lstlisting}[style=promptstyle]
{
  "name": "<FUNCTION_NAME>",
  "description": "<BRIEF_DESCRIPTION>",
  "parameters": {
    "<PARAM_NAME>": {
      "type": "<TYPE>",
      "range": [<MIN>, <MAX>],
      "description": "<Usage info or constraints>"
    },
    "<OPTIONAL_PARAM>": {
      "type": "string",
      "description": "<Optional note>"
    }
  },
  "required": ["<PARAM_NAME>"]
}
\end{lstlisting}
\end{tcolorbox}

\begin{table*}[!h]
\centering
\setlength{\tabcolsep}{3pt}
\renewcommand{\arraystretch}{0.95}

\begin{minipage}[t]{0.52\textwidth}
\vspace{0pt}
\centering
\captionof{table}{Supplemental results.}
\label{tab:moresults}
\resizebox{\linewidth}{!}{
\begin{tabular}{lcc}
\toprule
\multicolumn{1}{c}{\textbf{Method}}
& \multicolumn{1}{c}{\textbf{LIVEFB}} 
& \multicolumn{1}{c}{\textbf{TID}} \\
\midrule
\multicolumn{3}{l}{\cellcolor[HTML]{EFEFEF}\textit{VLM-based}} \\
Qwen3-VL-8B-Ins~\cite{yang2025qwen3technicalreport} 
& .357/.408 & .656/.706 \\
VisualQuality-R1~\cite{wu2025visualqualityr} 
& \underline{.423/.479} & \underline{.700/.752} \\
\midrule
\rowcolor[HTML]{E8F4FD}
\textit{Tool-IQA (Ours)} 
& \textbf{.468/.548} & \textbf{.733/.780} \\
\rowcolor[HTML]{F2F8FD}
\textit{Gains} 
& $.045/.069$ & $.033/.028$ \\
\bottomrule
\end{tabular}
}
\end{minipage}
\hfill
\begin{minipage}[t]{0.45\textwidth}
\vspace{0pt}
\centering
\captionof{table}{Ablation study \textbf{\textit{without training}}.}
\label{tab:ablation-tools-zeroshot}
\resizebox{\linewidth}{!}{
\begin{tabular}{lccccc}
\toprule
\multicolumn{1}{l}{\multirow{2}{*}{\textbf{Methods}}} 
& \multirow{2}{*}{\textbf{Mag.}} 
& \multirow{2}{*}{\textbf{Cor.}} 
& \multicolumn{3}{c}{\textbf{\textit{Scenarios}}} \\
\cmidrule(lr){4-6}
 & & & CLIVE & KonIQ & SPAQ \\ 
\midrule
\multicolumn{6}{l}{\textbf{SRCC}} \\
Qwen3-VL-8B-Ins   & $\times$ & $\times$ & 0.760 & \underline{0.729} & 0.856 \\
Tool-IQA   & $\checkmark$ & $\times$ & \underline{0.784} & 0.714 & \textbf{0.867} \\
\rowcolor[HTML]{E8F4FD}
Tool-IQA   & $\checkmark$ & $\checkmark$ & \textbf{0.794} & \textbf{0.732} & \underline{0.866} \\
\midrule
\multicolumn{6}{l}{\textbf{PLCC}} \\
Qwen3-VL-8B-Ins   & $\times$ & $\times$ & 0.772 & \underline{0.769} & 0.849 \\
Tool-IQA   & $\checkmark$ & $\times$ & \underline{0.802} & 0.758 & \underline{0.871} \\
\rowcolor[HTML]{E8F4FD}
Tool-IQA   & $\checkmark$ & $\checkmark$ & \textbf{0.828} & \textbf{0.781} & \textbf{0.872} \\
\bottomrule
\end{tabular}
}
\end{minipage}

\end{table*}

\begin{figure}[!h]
    \centering
    \includegraphics[width=0.49\linewidth]{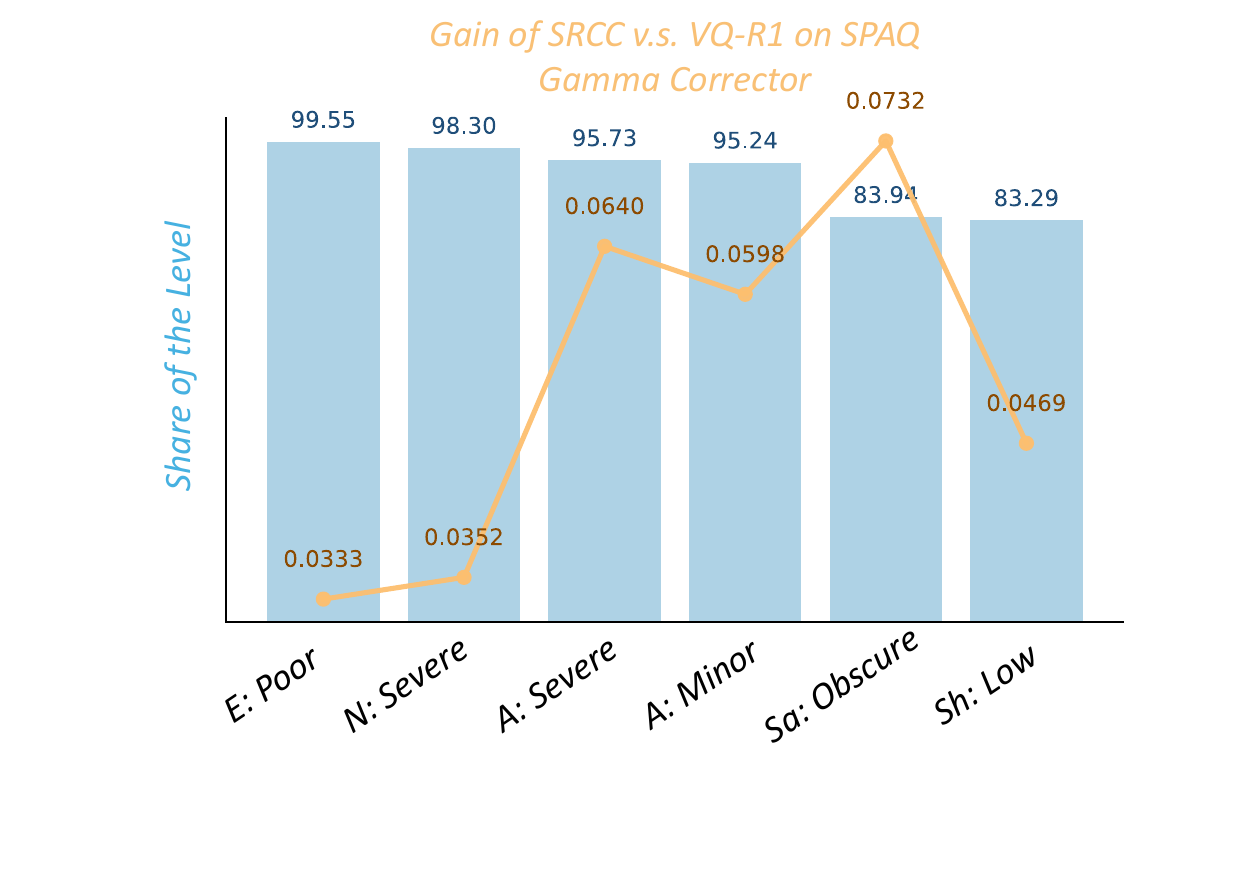}
    \includegraphics[width=0.49\linewidth]{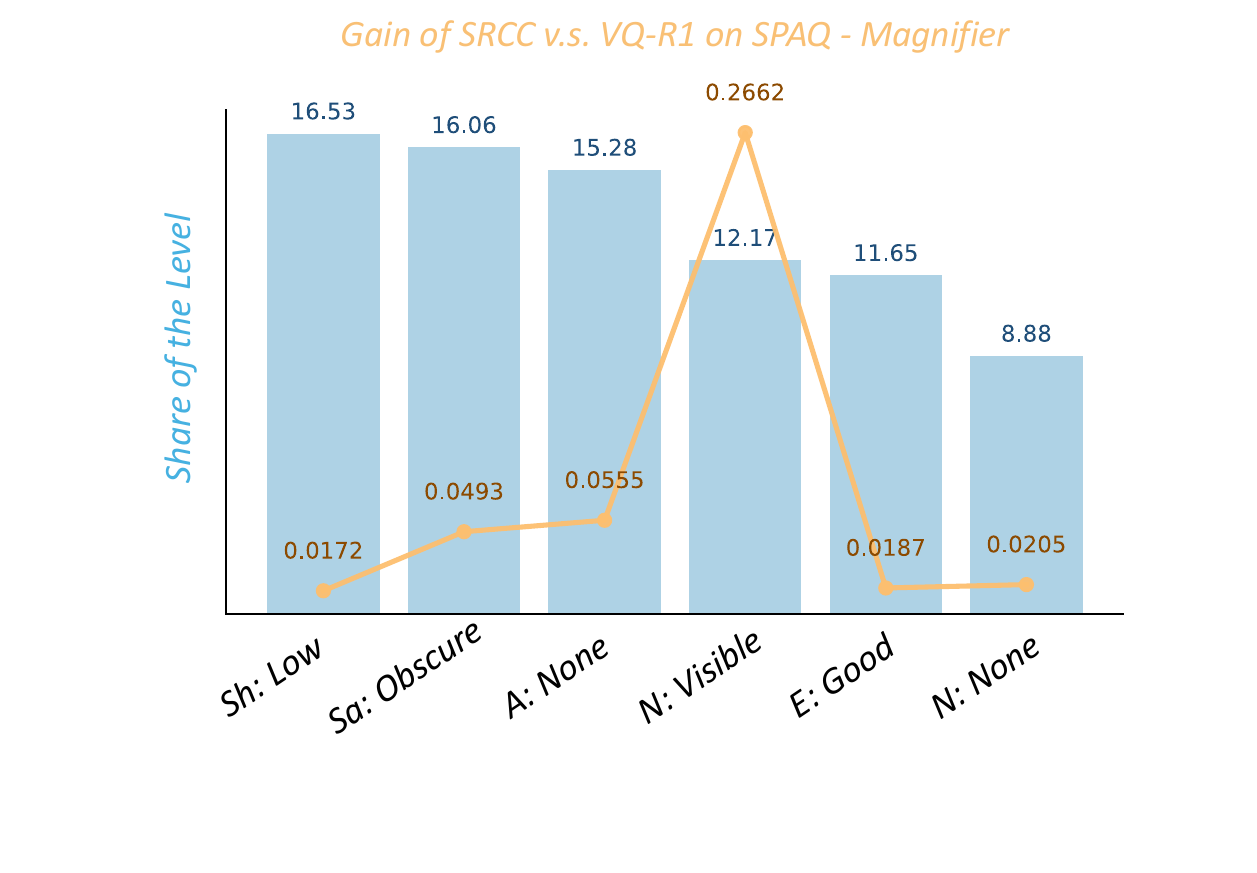}
    \caption{The distribution of tool usage and corresponding gains against induced rubric notes. Zoom in for better view.}
    \label{fig:rubric}
\end{figure}

\section{Supplemental Performance Study}

We further evaluate Tool-IQA on LIVEFB and TID to verify its generalization ability across diverse IQA benchmarks. As shown in Tab.~\ref{tab:moresults}, Tool-IQA consistently outperforms VisualQuality-R1 on these two datasets, demonstrating the effectiveness of tool-augmented visual quality assessment.

\section{Tool Effectiveness Study}
We supplement more ablation studies in Tab.~\ref{tab:ablation-tools-zeroshot}. Robust gains are also observed under the \textit{w/o training} setting, indicating that tool-assisted observations can benefit the VLM even without additional optimization. 
In addition, Fig.~\ref{fig:rubric} reveals distinct tool-usage patterns. We see that the model focuses on the six rubric attribute levels with the highest tool-utilization rates (by share within each attribute level), where the Gamma Corrector tool is predominantly invoked for lower-quality inputs to address visual impairments. Conversely, the magnifier function is utilized more for high-quality images to facilitate detailed inspection, a strategy that is fully consistent with the rubric-guided analysis.

\end{document}